\documentclass[letterpaper]{article} % DO NOT CHANGE THIS
\usepackage{aaai2026}  % DO NOT CHANGE THIS
\usepackage{times}  % DO NOT CHANGE THIS
\usepackage{helvet}  % DO NOT CHANGE THIS
\usepackage{courier}  % DO NOT CHANGE THIS
\usepackage[hyphens]{url}  % DO NOT CHANGE THIS
\usepackage{graphicx} % DO NOT CHANGE THIS
\usepackage{natbib}  % DO NOT CHANGE THIS AND DO NOT ADD ANY OPTIONS TO IT
\usepackage{caption} % DO NOT CHANGE THIS AND DO NOT ADD ANY OPTIONS TO IT
\usepackage{booktabs}

\usepackage{algorithm}
\usepackage{algorithmic}
\usepackage{amsmath}

\usepackage{newfloat}
\usepackage{listings}
\DeclareCaptionStyle{ruled}{labelfont=normalfont,labelsep=colon,strut=off} % DO NOT CHANGE THIS
\floatstyle{ruled}
\newfloat{listing}{tb}{lst}{}
\floatname{listing}{Listing}
\usepackage{siunitx}   % 用于 S 列
\usepackage{amssymb} % 推荐（包含更多数学符号）
\usepackage{amsthm}
\newtheorem{definition}{Definition} % 定义 definition 环境
\usepackage{multirow}

\title{Learning to Solve Weighted Maximum Satisfiability with a Co-Training Architecture}

\author{
    Written by AAAI Press Staff\textsuperscript{\rm 1}\thanks{With help from the AAAI Publications Committee.}\\
    AAAI Style Contributions by Pater Patel Schneider,
    Sunil Issar,\\
    J. Scott Penberthy,
    George Ferguson,
    Hans Guesgen,
    Francisco Cruz\equalcontrib,
    Marc Pujol-Gonzalez\equalcontrib
}
\author{
    Kaidi Wan\textsuperscript{\rm 1},
    Minghao Liu\textsuperscript{\rm 2},
    Yong Lai\textsuperscript{\rm 1}\textsuperscript{*}
}
\affiliations{
    \textsuperscript{\rm 1}Key Laboratory of Symbolic Computation and Knowledge Engineering of Ministry of Education, Jilin University, China \\
    \hspace*{1em} \texttt{wankd00@gmail.com}, \texttt{laiy@jlu.edu.cn} \\
    \textsuperscript{\rm 2}Department of Computer Science, University of Oxford, UK \\
    \hspace*{1em} \texttt{minghao.liu@cs.ox.ac.uk} \\
    \textsuperscript{*}Corresponding author
}

\usepackage{bibentry}
\begin{document}

\makeatletter
\def\@author{%
    Kaidi Wan\textsuperscript{\rm 1},
    Minghao Liu\textsuperscript{\rm 2},
    Yong Lai\textsuperscript{\rm 1}\textsuperscript{*}
}
\makeatother

\maketitle

\begin{abstract}
    We propose SplitGNN, a graph neural network (GNN)-based approach that learns to solve weighted maximum satisfiability (MaxSAT) problem.
    SplitGNN incorporates a co-training architecture consisting of supervised message passing mechanism and unsupervised solution boosting layer.
    A new graph representation called edge-splitting factor graph is proposed to provide more structural information for learning, which is based on spanning tree generation and edge classification.
    To improve the solutions on challenging and weighted instances, we implement a GPU-accelerated layer applying efficient score calculation and relaxation-based optimization.
    Experiments show that SplitGNN achieves $3\times$ faster convergence and better predictions compared with other GNN-based architectures.
    More notably, SplitGNN successfully finds solutions that outperform modern heuristic MaxSAT solvers on much larger and harder weighted MaxSAT benchmarks, and demonstrates exceptional generalization abilities on diverse structural instances.
\end{abstract}

% SplitGNN classifies edges into parent, child, and non-tree edges, with the option to further categorize non-tree edges as shallow or deep. 
%     This classification enables the identification of structural information, such as biconnectivity. 
%     This approach is applied to design customized SplitGNN for solving MaxSAT, achieving $\times 3$ faster convergence and better results. 
%     We believe this method has the potential to be applied to other graph-based problems as well.  
%     To improve the computation of variable scores and Best from Multiple Selections, we implement a GPU-accelerated, parallelized version using sparse matrices. 
%     This method enhances computational efficiency and scalability, enabling faster processing of large benchmarks. 
%     Finally, we improve previous approach for unsupervised MaxSAT solving and combine it with the two our methods to explore a hybrid search strategy that integrates both supervised and unsupervised relaxation techniques. 
%     Our method successfully finds optimal solutions for many MaxSAT benchmarks. 
\section{Introduction}
Boolean Satisfiability (SAT) is the first problem proven to be NP-complete \cite{cook1971complexity}. SAT and its optimization variant, Maximum Satisfiability (MaxSAT), are two core problems in computer science.
% which have been key technologies of the 21st century.
We have known that they are theoretically hard problems.
Although 2-SAT can be solved in polynomial time \cite{tarjan1972depth}, for general SAT and MaxSAT, unless P=NP, no polynomial-time algorithms are known. 
In particular, 2-MaxSAT is also NP-hard \cite{garey1974some}.
Given the fundamental role of SAT and MaxSAT in many complex formal reasoning challenges, with broad applications in artificial intelligence \cite{dimopoulos2012mu}, software engineering \cite{si2017maximum}, and operations research \cite{dixon2000combining}, there have been continuous efforts with the goal of efficiently solving more challenging problem instances in practice \cite{biere2021handbook,haberlandt2023effective}.
As said by Edmund Clarke, 2007 ACM Turing Award Recipient, ``Efficient SAT solving is a key technology for 21st century computer science''.
% Over the years, research on SAT and MaxSAT has led to significant practical and theoretical advances both in the field itself and in the areas it has influenced. 
%Specifically, a number of efficient MaxSAT solvers have been developed in the last few years, such as \cite{chu2023nuwls,zheng2024rethinking}.

However, traditional MaxSAT solving algorithms are based on enumerative search and intricate heuristics, making them awkward to automatically adjust strategies across different instances and not good at learning from solved instances.
% Although traditional methods for solving combinatorial problems, such as SAT and MaxSAT, are primarily based on symbolic search algorithms, there have consistently been efforts to address these challenges using deep learning approaches.
To bridge the gap in solvers in these aspects, there have consistently been efforts to address these problems with machine learning approaches.
One of the earliest attempts was the application of Hopfield networks \cite{hopfield1985neural} to the Traveling Salesman Problem (TSP). 
In recent years, deep learning has emerged as a powerful methodology for tackling complex combinatorial optimization problems \cite{cappart2023combinatorial}.
% , a category of problems known for their inherently intricate combinatorial structures.
Advances in deep representation learning have sparked initial efforts to tackle SAT \cite{selsam2018learning,duan2022augment} and MaxSAT \cite{liu2023can} with pure data-driven approaches, and Graph Neural Networks (GNNs) become the most widely used architecture.
Distinct from solvers that rely on enumerative search, the GNN-based methods are also called neural solvers.
They first transform input formulas into graphs, then utilize GNNs to learn structural characteristics through message passing and predict solutions to the problems.
Neural solvers have been found to demonstrate exceptional generalization abilities, where GNN models trained on simple instances are able to find even better solutions on much larger and harder instances compared with state-of-the-art traditional solvers.
Therefore, they are believed to be very promising approaches for solving challenging real-world instances.
% Mainstream Graph Neural Networks (GNNs) typically aggregate features of neighboring nodes in each layer and use these embeddings to form informative representations. 
% However, GNNs exhibit several limitations in common scenarios. 
% They struggle to capture long-range dependencies, making it difficult to handle long-range reasoning effectively \cite{dai2018learning}. 
% Distant signals are often overly compressed, leading to over-squashing issues that distort information flow \cite{alon2020bottleneck}. 
% In graphs characterized by heterophily, where connected nodes are often dissimilar, GNNs can propagate incorrect signals from dissimilar linked nodes \cite{zhu2020beyond}.   
% Furthermore, GNNs may fail to distinguish between two similar inputs, limiting their expressiveness in such scenarios \cite{xu2018powerful}. 
% Due to these shortcomings of GNN, some new methods have been designed. 
% For example, \cite{zhang2023rethinking} have designed a transformer-like architecture called GD-WL that combines the shortest distance and resistance distance to better recognize the graph biconnectivity. 
% These theoretical contributions provide a basis for us to further study machine learning methods on the MaxSAT problem. 

In this paper, we propose SplitGNN, a novel GNN-based approach that learns to solve the weighted MaxSAT problem. 
% SplitGNN classifies edges in a graph into parent, child, and non-tree edges, with an additional classification of non-tree edges into shallow or deep categories. 
To enrich the structural features that can be learned by GNNs, we represent input formulas as edge-splitting factor graphs, where edges are divided into four types based on a spanning tree.
% This edge classification enables the identification of structural features, such as biconnectivity, which are essential for optimizing solutions.
A message passing mechanism aligning with edge-splitting factor graph has been designed, which learns from training benchmarks in a supervised manner.
To further improve the quality of solutions, we introduce an unsupervised solution boosting (USB) layer, which implements efficient GPU-accelerated score calculation and relaxation-based optimization techniques.
This co-training architecture strengthens SplitGNN's ability to handle more challenging weighted MaxSAT instances.
%
% Our MaxSAT experiments demonstrate that SplitGNN outperforms existing models by achieving faster convergence and superior performance, especially on more complex problems. 
% Remarkably, even after model compression, SplitGNN continues to outperform models with a larger number of parameters. 
% These promising results suggest that SplitGNN could be adapted to various graph neural network architectures. 
% Furthermore, combining SplitGNN with the ResNet \cite{he2016deep} and Attention \cite{vaswani2017attention} mechanisms leads to further improvements. 
% Furthermore, we introduce an unsupervised MaxSAT solving method that combines parallel algorithms and relaxation techniques to further enhance the model's efficiency and scalability. 
%
Experiments on various synthetic benchmarks with different distributions demonstrate that SplitGNN outperforms the leading GNN-based models GMS-N and GMS-E \cite{liu2023can} on unweighted instances.
More importantly, SplitGNN is the first GNN-based method that successfully learns to solve weighted instances.
On the difficult weighted benchmarks with up to 2,000 variables and 20,000 clauses, SplitGNN can predict better solutions in a shorter time, even compared with state-of-the-art heuristic MaxSAT solvers Loandra \cite{berg2020loandra} and SATLike \cite{lei2021satlike}.
Additionally, testing on diverse structural MaxSAT instances highlights the strong generalization abilities of SplitGNN.
Our contributions can be summarized as follows:
\begin{itemize}
    % \raggedright % 强制左对齐
    % \item We propose a novel co-training GNN model called SplitGNN with edge classification mechanism, and use it to design customized version for MaxSAT to be solved with faster convergence and better results, and we believe that this method has the potential to be used in other problems on graphs.
    \item We propose a novel co-training GNN model called SplitGNN that can learn to solve both weighted and unweighted MaxSAT problem. To the best of our knowledge, this is the first end-to-end GNN-based method for weighted MaxSAT problem. 
    % SplitGNN combines a message passing mechanism and an unsupervised solution boosting layer to improve the solving efficiency.
    % \item For the computation of variable scores and Best from Multiple Selections, we implement a GPU-accelerated, parallelized version using sparse matrices. This method enhances computational efficiency and scalability, enabling faster processing of large benchmarks.
    \item We design edge-splitting factor graph, a new graph representation for CNF formulas. By generating a spanning tree, it provides richer structural information for GNNs to enhance their learning efficiency through message passing.
    % \item We improve our previous method for unsupervised solving of MaxSAT and combine it with our two previous contributions to explore a hybrid search, supervised and unsupervised, approach for solving MaxSAT. Our method finds optimal solutions on many benchmarks. 
    \item We present an effective GPU-accelerated unsupervised solution boosting layer, which combines score calculation module and relaxation-based optimization to consistently improve the quality of solutions.
    \item Comprehensive experiments on diverse benchmarks show that SplitGNN predicts better solutions compared with other GNN-based methods and heuristic solvers, including those large and challenging weighted MaxSAT instances.
\end{itemize}
\section{Preliminaries}

\subsection{MaxSAT} 
Given a Boolean variable set $V = \{x_1, x_2, \ldots, x_n\}$, the corresponding literal set $L$ can be constructed as $L = \bigcup_{i=1}^n \{x_i, \neg x_i\}$ where each element in $L$ represents either variable itself or variable negation. 
A clause $c_i$ is a disjunction (logical OR) of literals, formally expressed as: $c_i = l_{i1} \lor l_{i2} \lor \cdots \lor l_{ik}$ where $k$ denotes the \textit{length} of clause $c_i$.
A Conjunctive Normal Form (CNF) formula $\mathcal{F}$ consists of the conjunction (logical AND) of $m$ clauses: $\mathcal{F} = c_1 \land c_2 \land \cdots \land c_m$ where $m$ represents the total number of clauses in $\mathcal{F}$.
An illustrative CNF formula might be $(x_1 \vee x_2) \wedge (x_2 \vee \neg{x_3}) \wedge (x_1 \vee \neg{x_2} \vee x_3)$.
% CNF formulas can represent any propositional logic statement, and they can be efficiently generated from other forms using methods like Tseitin encoding \cite{tseitin1983complexity}.
An assignment is a mapping that assigns a Boolean value ($\mathit{true}$ or $\mathit{false}$) to each variable. 
Given an assignment $\alpha$, a clause c is satisfied if at least one literal in c is $\mathit{true}$, otherwise c is falsified.
The Boolean Satisfiability (SAT) problem involves determining whether there exists an assignment to the variables such that all clauses are satisfied.

The Maximum Satisfiability (MaxSAT) problem extends SAT by finding an assignment that maximizes the number of satisfied clauses, even if not all clauses can be satisfied simultaneously.
MaxSAT is especially important when dealing with over-constrained systems, where finding a solution that satisfies all constraints may not be feasible, but we aim to satisfy them as many as possible.
% The partial MaxSAT (PMS) problem is a variant of MaxSAT: In a partial CNF formula, clauses are divided into hard ones and soft ones, and the task is to find an assignment that satisfies all hard clauses and maximizes the number of satisfied soft clauses. 
Weighted MaxSAT problem is a more general variant of MaxSAT, where each clause is assigned a positive integer as weight, and the goal is to find an assignment that maximizes the sum of weights of satisfied clauses.
Another crucial variant, partial weighted MaxSAT problem, can be reduced to weighted MaxSAT by setting sufficiently large weight to hard clauses.
% Both SAT and MaxSAT are classified as NP-hard problems, indicating that no known algorithm can solve them in polynomial time for all cases unless P equals NP. However, these problems have significant practical applications, leading to the development of various solving techniques and heuristics that can handle large instances effectively in practice. Techniques originally developed for SAT, such as efficient data structures and strategies for selecting variables, have been adapted for use in MaxSAT solvers, and advancements in MaxSAT have also influenced the evolution of SAT-solving technologies.

% \subsection{Tarjan Algorithm}

\subsection{Graph Neural Networks}
Graph Neural Networks (GNNs) have emerged as a powerful architecture in deep learning.
Unlike traditional deep learning models, such as Convolutional Neural Networks (CNNs) and Recurrent Neural Networks (RNNs), which are designed for Euclidean data (like images or sequences), GNNs are specifically designed to handle non-Euclidean data, where relationships between data points are defined by the structure of graph.

%Graphs are ubiquitous in various fields, including social networks, recommendation systems, bioinformatics, computer vision and natural science. 
Most popular GNN models adopt the message passing process, where a node's embedding is updated iteratively by aggregating information from its neighbors. 
The operation at the $k$-th iteration (or layer) of GNNs can be formalized as:
\begin{equation}
    \begin{aligned} 
        s_v^{(k)} &= \text{AGG}^{(k)}\left(\left\{h_u^{(k-1)} \mid u \in \mathcal{N}(v)\right\}\right), \\
        h_v^{(k)} &= \text{UPD}^{(k)}\left(h_v^{(k-1)}, s_v^{(k)}\right),
    \end{aligned}
\end{equation}
where \(h_v^{(k)}\) represents the embedding vector of node \(v\) after the \(k\)-th iteration. In the aggregation (or messaging) step, a message \(s_v^{(k)}\) is generated for each node \(v\) by collecting the embeddings from its neighbors \(\mathcal{N}(v)\).
Next, in the update step, \(h_v^{(k)}\) is obtained by combining \(s_v^{(k)}\) and the previous embedding \(h_v^{(k-1)}\).
After $T$ iterations, the final embedding $h_v^{(T)}$ is obtained for each node $v$.
% If the task GNNs need to predict is at the edge level, such as link prediction, an embedding vector of the edge should be generated from two vertices at the ends of the edge.
% For tasks at the subgraph or entire graph level, an embedding vector for the graph will be derived by integrating the embeddings of its constituent vertices.
% Modern GNN variants employ diverse choices for the aggregation and update functions, such as concatenation, summation, max-pooling, and mean-pooling.
% Some recent research has attempted to combine graph neural networks with attention mechanism \cite{velivckovic2017graph,yun2019graph,wu2024simplifying}.
A more detailed and comprehensive review of GNNs is available in \cite{corso2024graph}. 

\subsection{GNN-based MaxSAT Solving}
GMS is a GNN-based method proposed in \cite{liu2023can} for learning to solve unweighted MaxSAT problem.
It contains two very similar GNN models, called GMS-N and GMS-E, based on two variants of factor graphs.
% The first one is node-splitting, which represents the two relevant literals as two nodes. 
% The other one is edge-splitting, which contains two types of edges, connecting the clauses with positive and negative literals separately. 
% Subsequently, the authors implement Graph Neural Networks (GNNs) on the two graph structures, employing Multi-Layer Perceptrons (MLPs) for neighbor aggregation and Long Short-Term Memory (LSTM) networks for node state updating, to perform numerical prediction of variable assignment. 
GMS models employ Multi-Layer Perceptrons (MLPs) with mean-pooling for neighbor aggregation and Long Short-Term Memory (LSTM) networks for updating node embeddings.
According to experimental results, GMS models outperform state-of-the-art heuristic solvers on two groups of MaxSAT instances with up to 1,600 variables, which is the most effective end-to-end GNN-based method on this task.
However, GMS models do not support weighted MaxSAT problem, a more general and widely applicable task in real-world scenarios.

\section{Related Work}
Currently, in addition to single-core methods for finding optimal solutions to combinatorial optimization problems related to MaxSAT, many new approaches have emerged \cite{berg2024maxsat}.   
\cite{chu2023nuwls} and \cite{zheng2024rethinking} are outstanding local search technique to solve MaxSAT that can guide the direction of the solution. 
% \cite{zhao2024distributed} presents a distributed SMT solving approach based on dynamic partitioning at the variable level. 

We focus on the integration of deep learning with satifiability. %combinatorial optimization techniques and methods. 
% \cite{silver2016mastering,schrittwieser2020mastering} present groundbreaking approaches that combine deep neural networks with Monte Carlo tree search to achieve superhuman performance in the game of Go.  
NeuroSAT \cite{selsam2018learning} is a pioneering work which shows that GNNs can learn to predict the satisfiability of SAT problems, and leads to a series of improvements.
% \cite{balunovic2018learning} presents a new approach that combines learning and synthesis techniques to automatically find effective SMT solving strategies.
\cite{yolcu2019learning} learns local search heuristics through deep reinforcement learning.  
\cite{kurin2020can} focuses on learning the MiniSat solver heuristics. 
% \cite{shi2022satformer} introduces SATformer, a novel Transformer-based approach that focuses on identifying unsatisfiable subproblems by modeling clause interactions. 
% \cite{li2022nsnet} presents NSNet, a general neural probabilistic framework that models satisfiability problems as probabilistic inference, achieving results in solving SAT and \#SAT problems through marginal inference and approximate model counting.   
\cite{duan2022augment} uses Label-Preserving Augmentations in Graph Contrastive Learning for solving SAT problems. 
\cite{li2023hardsatgen} reconstructs CNF formula using node Splitting and merging operations of graphs.
% \cite{liu2023can} investigates the capability of Graph Neural Networks (GNNs) to learn and solve Maximum Satisfiability (MaxSAT) problems, demonstrating their potential through experimental evaluations and providing a theoretical explanation based on algorithmic alignment theory. 
% \cite{shirokikh2023machine} proposes a strategy that combines machine learning with classical heuristics for SAT solving, using a trained model for initial steps and then transitioning to traditional heuristics, while also introducing a modified Graph-Q-SAT framework tailored for SAT problems derived from other domains like open shop scheduling. 
\cite{zhang2024grass} utilizes GNNs and global pooling to extract features for selecting sat solver.  
% \cite{wang2024grounding} introduces SMTLayer, a method for integrating Satisfiability Modulo Theories (SMT) solvers into deep neural network layers, which leverages symbolic knowledge during training and inference to improve model performance in logical reasoning tasks. 
% \cite{warde2023solving} introduces RbmSAT, a novel algorithm that leverages Restricted Boltzmann Machines (RBMs) and matrix multiplication on neural network accelerators (TPUs/GPUs) to efficiently solve MaxSAT problems. 
% By combining parallel Gibbs sampling and a unit propagation-based heuristic, it outperforms traditional CPU-based solvers in part of unweighted benchmark evaluations. 
\cite{hosny2024torchmsat} uses GPU acceleration and relaxation operations to solve the MaxSAT in an unsupervised manner. 

% \cite{lei2022cashwmaxsat} 
\section{SplitGNN for MaxSAT Solving}

\paragraph{Intuition of SplitGNN}
% SplitGNN is similar to Mixture of Experts, but it does not use parameters to determine which expert to use but rather the structure of the graph. 
Spanning tree network allow vertices on two identical connected blocks to perceive basic path connections, while non-spanning tree network allow connected vertices to perceive more information passed by different paths. 
Using spanning tree decomposition, the edges on the tree can form a structure similar to assignment propagation, which can lead to an assignment that satisfies many subgroups of clauses. 
Meanwhile, non-tree edges represent possible conflicts, through which information can be propagated from two directions to the nodes within a substructure, helping the model to analyze and resolve conflicts. 

\subsection{Edge-Splitting Factor Graph}

\begin{figure}[b]
    \centering
    \includegraphics[width=0.63\linewidth]{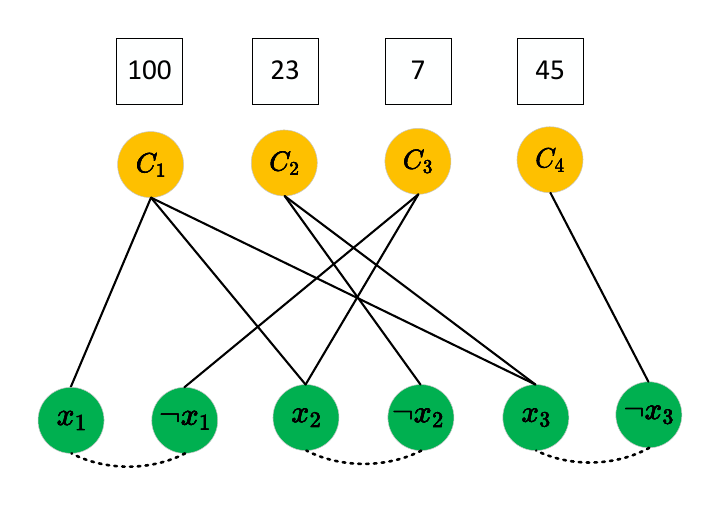}  % 使用 \includegraphics 加载 .pdf 文件
    \caption{A Factor graph to represent the WCNF formula $(x_1 \lor x_2 \lor x_3) \land (\neg{x_2} \lor x_3) \land (\neg x_1 \lor x_2) \land \neg x_3$ with 3 variables and 4 clauses, where the weights of the clauses are $100, 23, 7, 45$, respectively.
    This formula is unsatisfiable. When $x_1=1,x_2=0,x_3=1$, the satisfied clauses have the largest weights, which is an optimal solution of this problem.
    % Similar formulas and their factor graphs can be solved by graph neural networks and unsupervised methods that we design later.
    }
    \label{fig:CNF_Graph}
\end{figure}

Following the convention of previous works such as \cite{selsam2018learning,liu2023can}, we represent CNF formulas as factor graphs.
Figure \ref{fig:CNF_Graph} illustrates an example of how to transform a CNF formula into a factor graph. A factor graph is a triple $G=\langle V_L, V_C, E \rangle$, where $V_L$ is the set of literal nodes, $V_C$ is the set of clause nodes, and $E$ is the set of edges. A literal $l_i \in V_L$ and a clause $C_j \in V_C$ are connected by an undirected edge $(l_i,C_j) \in E$ if and only if $C_j$ contains $l_i$.
Moreover, each literal $x_i$ and its negation $\neg{x_i}$ are also connected by a dashed edge, which means that there is no actual edges between them, but their embeddings will be exchanged within GNNs.
% When dashed edges are not considered, the factor graph is bipartite which can only have even cycles. 
% After considering dashed edges, the whole graph may have more cycles as well as odd cycles.

Although factor graphs have been equivalent representation of CNF formulas, there are some negative impacts when GNNs learn from them.
On the one hand, when updating the embedding of a node in factor graph, all neighboring nodes' information is equally important, which differs from the domain knowledge in MaxSAT solving where a part of the formula, such as backbones \cite{kilby2005backbones} and backdoors \cite{williams2003backdoors}, are more critical for the solution.
On the other hand, in a factor graph, node embeddings may propagate through many loops, causing GNNs to carry a lot of redundant information while learning the message passing functions, thus decreasing the effectiveness of learning.
To optimize the learning efficiency of GNNs, we propose the edge-splitting factor graph, which incorporates richer structural information through the computation of spanning trees.
We formally define the edge-splitting factor graph as follows:
\begin{definition}[Edge-splitting factor graph]
\label{def:esfg}
Given a factor graph $G=\langle V_L, V_C, E \rangle$, and a spanning tree $T=\langle V_L,V_C,E_T,v_{root} \rangle$ of $G$, an edge-splitting factor graph is a tuple $G_{ES}=\langle V_L,V_C,E_{parent},E_{child},E_{ntup},E_{ntdown} \rangle$. $E_{parent},E_{child},E_{ntup},E_{ntdown}$ are disjoint sets of directed edges and $E_{parent} \cup E_{child} \cup E_{ntup} \cup E_{ntdown}=E$. Specifically,
\begin{itemize}
    \item $E_{parent}$ {\rm \textsf{(Parent edges)}}: A directed edge $(x,y) \in E_{parent}$ iff $(x,y) \in E_T$ and $d(x) < d(y)$, where $d(u)$ is the distance between node $u$ and $v_{root}$ on $T$.
    \item $E_{child}$ {\rm \textsf{(Child edges)}}: A directed edge $(x,y) \in E_{child}$ iff $(x,y) \in E_T$ and $d(x) > d(y)$.
    \item $E_{ntup}$ {\rm \textsf{(Non-tree up edges)}}: A directed edge $(x,y) \in E_{ntup}$ iff $(x,y) \notin E_T$ and $d(x) > d(y)$.
    \item $E_{ntdown}$ {\rm \textsf{(Non-tree down edges)}}: A directed edge $(x,y) \in E_{ntdown}$ iff $(x,y) \notin E_T$ and $d(x) < d(y)$.
\end{itemize}
\end{definition}

\begin{figure}[htb]
    \centering
    % \raggedleft 
    % \hspace{15cm}
    \includegraphics[width=0.85\linewidth]{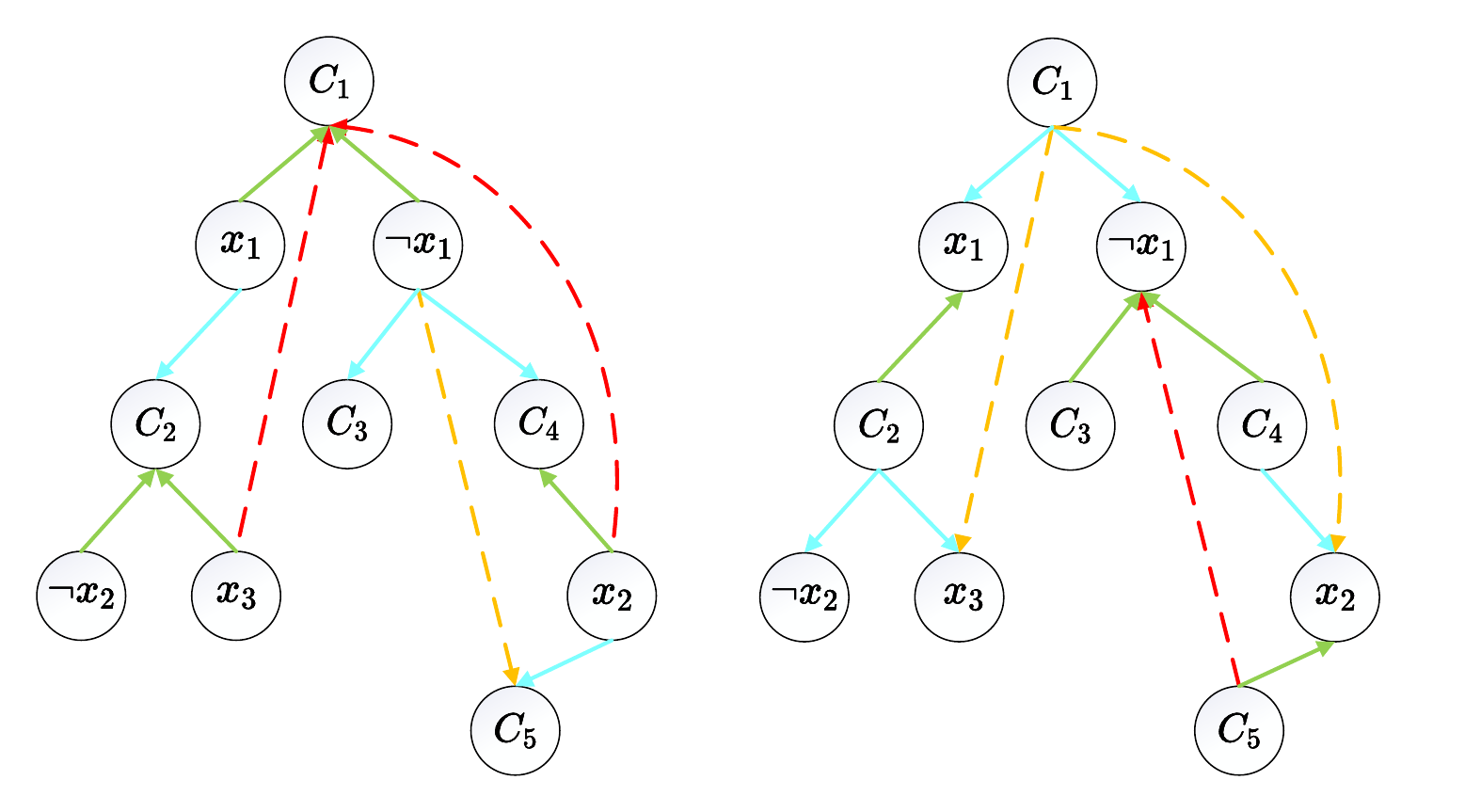}  % 使用 \includegraphics 加载 .pdf 文件
    \caption{A round of message passing in SplitGNN. The embeddings are propagated through four different aggregation functions represented by four types of edges: cyan for parent edges, green for child edges, red for non-tree up edges, and gold for non-tree down edges. Left: The embeddings of clauses are updated by the embeddings of literals. Right: The embeddings of literals are updated by the embeddings of clauses.}
    \label{fig:SplitGNN}
\end{figure}

% in some way by selecting some vertex as the root and partitioning the edges connected to any vertex V into four types of edges with V as the subject, edges connected to its father are called Parent edges, edges connected to a number of its children are called child edges, and edges that do not lie on T that come from deeper depths are called non-tree up edges (ntup), and edges that come from shallower are called non-tree down edges(ntdown).

We suggest that the proposed edge-splitting factor graph can contribute to improving the learning efficiency of GNNs from the following aspects.
First, by splitting the edges according to a spanning tree, we strengthen the heterogeneity of the graph while maintaining the connectivity among all nodes, which may aid GNNs in learning effective patterns from partial formulas.
Second, since different types of edges are handled by separate aggregation functions in GNNs as described later, each aggregation function does not accept embeddings of the same node through loops, which reduces the impact of redundant data on learning.
Third, by distinguishing edges with different directions based on the generated spanning tree, each node can grasp a better awareness of its global structural information in the graph.

\iffalse
Given a factor graph, we generate its spanning tree as described below.
We select the root node to be the one that minimizes the longest distance to the other nodes in each connected component. 
Then, we use a breadth-first search to extend the spanning tree
%by finding the node that is not traversed and is closest to the root    [or depth-first search]
, and adding the edge into the tree. 
Finally, we split all edges in factor graph into the four types according to Definition \ref{def:esfg}, which are accepted by our SplitGNN model.
\fi

\subsection{Message Passing of SplitGNN}

\begin{figure*}[tb]
    \centering
    \includegraphics[width=0.62\linewidth]{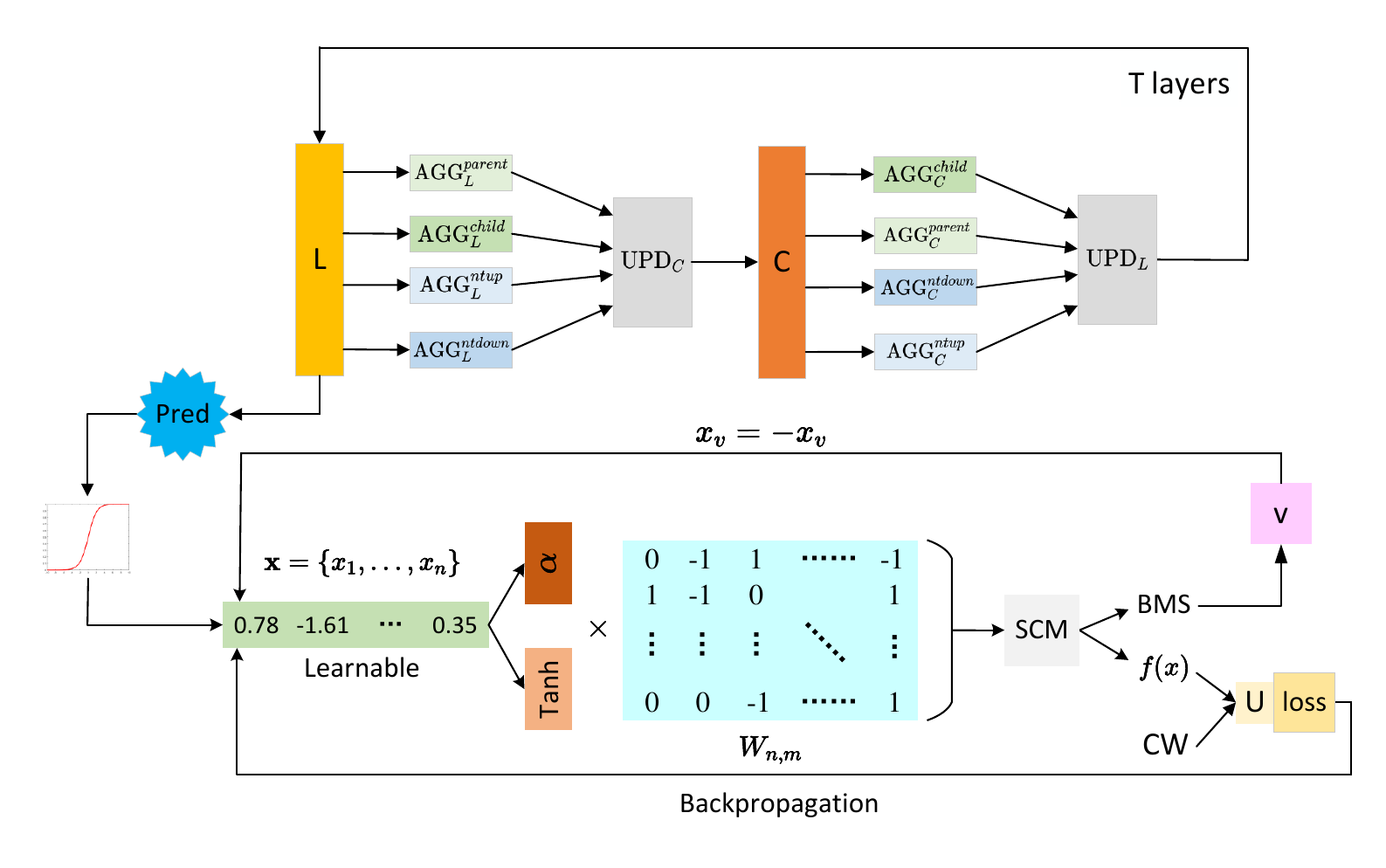}  % 使用 \includegraphics 加载 .pdf 文件
    \caption{The architecture of SplitGNN. The embeddings of literals are first updated through message passing for $T$ layers. Next, the prediction of variables is fed into an unsupervised solution boosting (USB) layer, where GPU-accelerated score calculation and relaxation-based optimization techniques are employed to generate better solutions.
    The better solution found can update the slack variable values and the parameters of the message passing to achieve transfer learning. }
    \label{fig:TSMaxSAT}
\end{figure*}

% The input vector \(x = \{x_1, x_2, \dots, x_n\}\) is processed through a learnable transformation, followed by a Tanh activation function. The result is multiplied by a matrix \(W_{n,m}\) representing the interaction between variables. The processed output is passed through the Parallel Scores Calculating (PSC) and BMS blocks, generating the output \(v\), which is then fed into the function \(f(x)\) for backpropagation. This architecture leverages SplitGNN for efficient MaxSAT problem solving.

Based on the edge-splitting factor graph, we propose SplitGNN, a GNN model that learns to solve the weighted MaxSAT problem.
Similar to the general architectures of GNNs, our SplitGNN propagates structural information of the graph between nodes through a message passing mechanism.
% SplitGNN traverses the graph with methods such as depth-first search or breadth-first search to obtain spanning tree and non-tree edges. 
% Then the Splitting continues by Splitting the edges on the spanning tree into parent and child edges, and Splitting the non-tree edges into non-tree up edges and non-tree down edges. 
% For each round of propagation, different edges connected by vertices are used with different aggregation functions before being merged to get the embedding vectors of the points in the current round.
Compared with the existing GNN models, for each round of propagation, SplitGNN uses different aggregation functions for different types of edges before the embeddings are merged.
This allows each node to better understand its structural information in the graph, which leads to the improvement of learning accuracy.
Figure \ref{fig:SplitGNN} demonstrates a round of message passing in SplitGNN.

Algorithm \ref{alg:Split_gms} describes the message passing machanism of SplitGNN in detail. We represent the four types of edges in an edge-splitting factor graph as four adjacency matrices: $M_{parent}$, $M_{child}$, $M_{ntup}$, and $M_{ntdown}$. 
Each of these matrices has a shape of \(|V_C| \times |V_L|\)\footnote{Since the non-zero entries in these matrices are few, we store them as sparse matrices. The actual size of storage space grows linearly with the total number of literals appearing in all clauses.}. 
The sparse neighbor order matrix and the aggregated node embeddings are multiplied to get the node embedding expression adjacent to each node, and then we use the edge classification mechanism of SplitGNN to classify the edges into the above four kinds to get more news about the graph and itself. 
The four types of features obtained are further aggregated in order to get the embedding of each clause node of the current layer. 
Four types of edges form four subgraphs, the edges on each subgraph are the same kind of edges. Message passing is used for each type of subgraph, and then through a multi-layer perceptron aggregation, from four times the dimension to the previous dimension. 
Attention \cite{vaswani2017attention} is used to aggregate literal embeddings before updating with two heads since two kinds of literals for each variable.
The update functions for both clause and literal embeddings are implemented using LSTM \cite{hochreiter1997long}. 
Since the edge-splitting factor graph is a bipartite graph, we transpose the four matrices for multiplication to get the embeddings of literal nodes.
% Since SplitGNN supports weighted MaxSAT solving, 
And we encode the weights of clauses into a matrix $CW$ and multiply it with the $L2C$ features within the update function.

\begin{algorithm}[!htb]
    \caption{Message Passing of SplitGNN}
    \label{alg:Split_gms}
    \raggedright % 强制左对齐
    \textbf{Input}: $G_{ES} = \langle V_L, V_C, E_{parent}, E_{child}, E_{ntup}, E_{ntdown} \rangle$ \\
    \textbf{Output}: The assignment prediction of variables $P$ \\
    \begin{algorithmic}[1] %[1] enables line numbers
        \STATE Build adjacency matrices $M_{parent}$, $M_{child}$, $M_{ntup}$, and $M_{ntdown}$ from spanning tree;
        \STATE Initialize $L^{(0)} \in \mathbb{R}^{|V_L| \times d}$;
        \STATE Initialize $C^{(0)} \in \mathbb{R}^{|V_C| \times d}$;
        \FOR{$k = 1$ to $T$}

            \STATE ${L2C^{(k)}} = \text{AGG}_L\big($
            \STATE \quad\quad $M_{parent} \cdot \text{AGG}_L^{parent}(L^{(k-1)}),$
            \STATE \quad\quad $M_{child} \cdot \text{AGG}_L^{child}(L^{(k-1)}),$
            \STATE \quad\quad $M_{ntup} \cdot \text{AGG}_L^{ntup}(L^{(k-1)})$,
            \STATE \quad\quad $M_{ntdown} \cdot \text{AGG}_L^{ntdown}(L^{(k-1)}) \big)$;
        
            \STATE $C^{(k)} \gets \text{UPD}_C\big(C^{(k-1)}, CW \cdot L2C^{(k)}\big)$;

            \STATE ${C2L^{(k)}} = \text{AGG}_C\big($
            \STATE \quad\quad $(M_{child})^\top  \cdot \text{AGG}_C^{parent}(C^{(k-1)}),$
            \STATE \quad\quad $(M_{parent})^\top \cdot \text{AGG}_C^{child}(C^{(k-1)}),$
            \STATE \quad\quad $(M_{ntup})^\top \cdot \text{AGG}_C^{ntdown}(C^{(k-1)})$,
            \STATE \quad\quad $(M_{ntdown})^\top \cdot \text{AGG}_C^{ntup}(C^{(k-1)}) \big)$; 
            
            \STATE $L^{(k)} \gets \text{UPD}_L\big(L^{(k-1)}, \neg L^{(k-1)}, C2L^{(k)}\big)$;
        \ENDFOR
        \STATE $P \in \mathbb{R}^{|V_L|/2} \gets \text{Sigmoid}\big(\text{PRED}_L(L^{(T)}, \neg L^{(T)})\big)$;
        %\STATE $\Phi \gets \text{Round}\big(\text{Sigmoid}(L_{\text{pred}})\big)$.
        \STATE \textbf{return} $P$;
    \end{algorithmic}
\end{algorithm}

Finally, a fully-connected binary classifier is utilized to decode the embeddings of the positive and negative literals with respect to each variable into the assignment prediction.
The binary cross-entropy (BCE) loss is employed, which is expressed as: 
\begin{equation}
\text{BCE}(y, p) = -(y \log(p) + (1 - y) \log(1 - p)),
\end{equation}
where \(p \in [0, 1]\) represents the predicted probability of a variable being \textsf{True}, and \(y \in \{0, 1\}\) denotes the assignment of a variable derived from the optimal solution.
The overall loss for a problem instance is computed by averaging the losses of all variables, which should be minimized during training.

% \subsection{Parallel Scores Calculating}
\subsection{Unsupervised Solution Boosting Layer}

To further improve the ability of SplitGNN to solve challenging MaxSAT problems, we propose the unsupervised solution boosting (USB) layer, which aims to produce better solutions through an unsupervised approach.

% Algorithm \ref{alg:TSMaxSAT} describes the pipeline of the USB layer.
% It accepts the assignment prediction of variables $P$, the WCNF formula $\phi$, and the maximum iteration steps $L_{max}$ as input.
% (What does the algorithm do?) In each step, ...
We first run message passing to get a solution prediction (Algorithm \ref{alg:Split_gms}), and then calculate the variable scores according to Algorithm \ref{alg:Parallel Scores Solving}. 
Inspired from BMS \cite{cai2015balance} if we can obviously find a better solution, we greedily update it, otherwise we use a loss function to update the variables' slack prediction. At the same time, we change the parameters in the message passing to achieve the effect of migration learning.
After that, we go back to the beginning of the USB layer and loop through the above process.
Note that the USB layer is fully compatible with GPU, thus enabling efficient acceleration.

\paragraph*{Score Calculation with Sparse Matrix}
We first introduce the Score Calculation with Sparse Matrix (SCM) module, which aims to evaluate the quality of the current assignment.
% This module is important to improve the solutions effectively in an unsupervised manner.
Algorithm \ref{alg:Parallel Scores Solving} describes how this module works.
The matrix $W$ is fixed and represents a given MaxSAT instance in CNF. 
Rows correspond to boolean variables, while columns correspond to clauses. 
A value of $CW_j$ or $-CW_j$ is assigned if a variable $x_i$ or $\neg x_i$ appears in clause $C_j$, respectively; otherwise, it is 0. 
$W_{pos}$ equals the absolute value of $W$. 
And we get the unit matrix $W_{init}$ by turning the positive and negative numbers in $W$ into +1 and -1. %, respectively. 
$S$ is a tensor of size equal to the number of clauses, with the value of the i-th element the opposite of the number of clause literals. 
In general, equation $\alpha \times W_{init} == S + 2k$  can be used to determine which clauses are satisfied by exactly $k$ literals, but here we just compute the unsatisfied clauses and the clauses that satisfy exactly one literal. 
So we can calculate $Q = \frac{1}{2}(\alpha \times W_{init} - S)$ to know how many literals of each clause are true by the current assignment. 
In practice, we calculate the matrix $U$ to represent whether each clause is unsatisfiable, and the matrix $O$ to represent whether each clause is satisfied by only one literal. 
$U$ and $W_{pos}$ are multiplied element-wise to obtain $M_u$ can know the positive score that each clause can provide after flipping each variable, followed by summation to obtain the positive score vector $Add$. 
Similarly $O$ and $W$ multiply by elements to extract the clauses where only one literal is satisfied, and then multiply with $\alpha$ to get $M_o$ by elements to obtain the satisfiability of the literals of these clauses.
We extract $(val, clause\_id)$ from $M_o$, use scatter to find $(max\_pos, max\_val)$ (weights and positions of satisfied literals), accumulate them into $Del$, and compute scores as $Add - Del$. 
This handles both weighted and unweighted MaxSAT.

\iffalse
Get the weights and clause indexes ($val, clause\_id$) by the sparse matrix $M_o$, then use scatter to find the maximum weights and corresponding indexes, ie, the indexes and weights ($max\_pos, max\_val$) of the literals that are satisfied.
Then we add the corresponding weights at the corresponding positions to the vector $Del$.
Finally, $Add$ is subtracted from $Del$ to obtain the scores of the variables. 
% $(i, j)$ denotes that clause $C_j$ has a positive or negative score contribution to $l_i$. 
This algorithm can handle both unweighted and weighted MaxSAT instances. 
\fi

\begin{algorithm}[htb]
    \caption{Score Calculation with Sparse Matrix}
    \label{alg:Parallel Scores Solving}
    \raggedright % 强制左对齐
    \textbf{Input}: The current assignment $\alpha$, the weighted CNF formula $\varphi$, matrix $W$\\
    \textbf{Output}: Scores of variables \\
    \begin{algorithmic}[1] %[1] enables line numbers
        \STATE Initialize $S$ where $S_i = -|C_i|$;
        \STATE Construct $W_{unit}$, $W_{pos}$ from $\varphi$;
        \STATE $U = (\alpha \times W_{unit} == S),\ \ O = (\alpha \times W_{unit} == S + 2)$;
        \STATE $M_{u} = U \circ W_{pos}, \ \ M_{o} = O \circ W \circ \alpha^T$;
        \STATE $Add$ = sum$(M_{u}, dim1)$;
        \STATE Get\ $val, clause\_id$ from $M_{o}$;
        \STATE $max\_pos, max\_val$ = scatter\_max$(val, clause\_id)$;
        \STATE $Del$ = scatter\_add$(max\_pos, max\_val)$;
        \STATE $Score = Add - Del$;
        \STATE \textbf{return} $Score$;
    \end{algorithmic}
\end{algorithm}

% \subsubsection{Hybrid Torch and Search}
% \paragraph*{Hybrid SplitGNN and Unsurpervised Search (Relaxation-based Optimization?)}
\paragraph*{Relaxation-based Optimization}
% To effectively address MaxSAT problems, SplitGNN incorporates an unsupervised solving approach that combines sparse matrix multiplication with relaxation techniques. 
% This methodology enhances both the efficiency and effectiveness of the solver.

\cite{hosny2024torchmsat} use the sum of relaxed literal values in unsatisfied clauses as the loss function. Inspired by this, we also relax the variables into real-valued vector $x$, where $x_i < 0$ corresponds to \textsf{False} and $x_i > 0$ to \textsf{True}.
But our approach has several differences with the previous work. 
First, we introduce the absolute value of sampling on the normal distribution to determine the size of $x$, and message passing to determine the positive and negative sign of $x$. 
This allows us to find the better solutions quickly. 
Then, we introduce Algorithm \ref{alg:Parallel Scores Solving} to determine if the current $x$ can be optimized with simple greedy optimization similar to BMS, if so then flip $x$, otherwise update $x$ with our loss function. 
Our loss function is designed as follows: We compute the function $f(x) = \tau \cdot \frac{\tanh(x)}{score^2+\epsilon} \times W_{unit}$, where $\tau$ is the temperature coefficient, $\tanh(x)$ is the normalized slack prediction, $\epsilon$ is a small constant. For the next step, we take the unsat  positions in $f(x)$ with the zero vector to compute the mean squared error. 
% In order to cope with unequal length clauses and normalization, divide by $S$.
Essentially, the slack prediction is used to calculate exactly how unsatisfiable each unsatisfied clause is and to move it towards satisfaction.
Dividing by the square of the score is affecting the weight of each variable corresponding to the slack value, as the variable scores are all non-positive at this point, which allows variables with larger scores to be prioritized for modification with larger weights.
At the same time, we consider how to optimize clause weights here, which addresses the inability of GMS models and significantly improves the results.
The flexibility of our method enables GNNs to explore a broader solution space more efficiently, leading to faster convergence to near-optimal solutions. 

\section{Experiments}
In this section, we make a comprehensive evaluation on the performance of SplitGNN.
We train SplitGNN on various sets of (weighted) MaxSAT benchmarks and compare it with GNN-based learning methods and also state-of-the-art heuristic MaxSAT solvers.
Moreover, we present that both the proposed message passing mechanism and the USB layer are beneficial for improving the solutions.

\subsection{Benchmarks} 
In the context of deep learning techniques, it is necessary to provide a large amount of data with the same distribution for training.
However, in the public MaxSAT benchmarks, each group of instances typically contains from a few to several dozen, which is far insufficient for the models to learn effective strategies.
Therefore, like other learning-based approaches, we experiment using diverse sets of synthetic benchmarks.
Specifically, we employ the following four instance generators:
% Uniform (UF) \cite{mitchell1992hard}, SR \cite{selsam2018learning}, Power Random 3SAT (PR) \cite{ansotegui2008random}, Double Power (DP) and Popularity Similarity (PS) \cite{giraldez2017locality,giraldez2021popularity}.
Uniform (UF) \cite{mitchell1992hard}, Power-Law (PL) \cite{ansotegui2008random}, Popularity Similarity (PS) and Double Power (DP) \cite{giraldez2017locality,giraldez2021popularity}.
UF and PL are synthetic generators with uniform and non-uniform clause selection probabilities, respectively.
PS and DP are pseudo-industrial generators producing instances that mimic real-world problems.
For unweighted MaxSAT problem, we denote a generated benchmark as ``AA($k$,$n$,$m$)'', where AA represents the generator, $k$ represents the (average) size of each clause, $n$ represents the number of variables, and $m$ represents the number of clauses.
For weighted MaxSAT problem, the weight of each clause is randomly set from 1 to 100, and we denote a generated benchmark as ``WAA($k$,$n$,$m$)'', where the letter `W' represents `weighted'.
Unless explicitly stated, on a benchmark, we train each model with 18,000 instances and use 2,000 independent instances for testing.
% The test set consisted of 2000 instances.

\subsection{Experimental Results} 

\paragraph*{Comparison with GNN-based Models}

\begin{figure*}[tb]
    \centering 
    \includegraphics[width=0.245\linewidth]{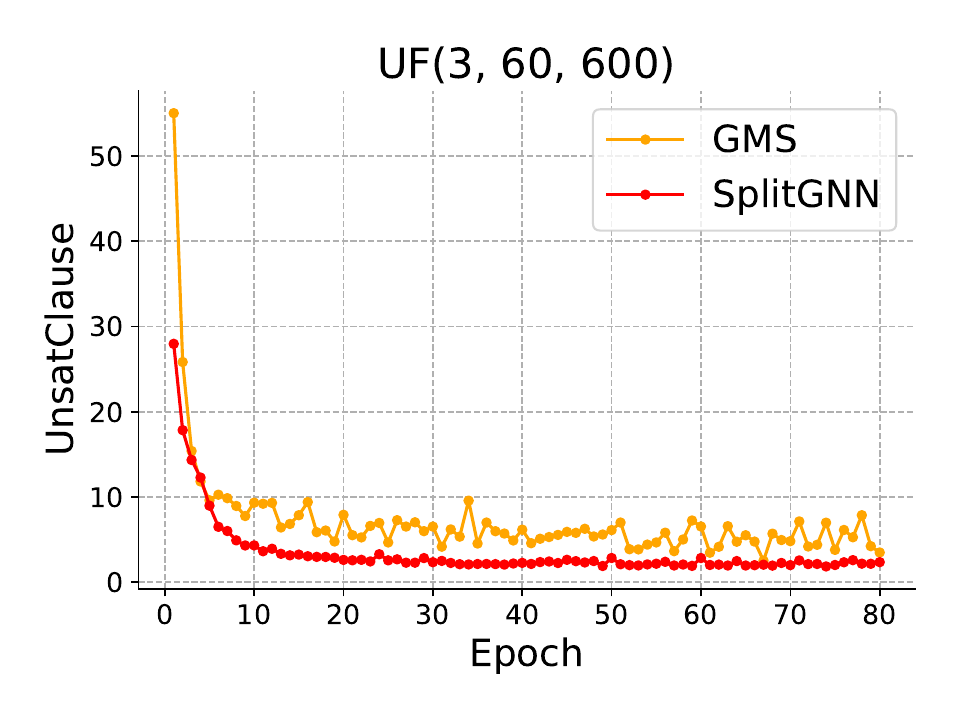}
    \includegraphics[width=0.245\linewidth]{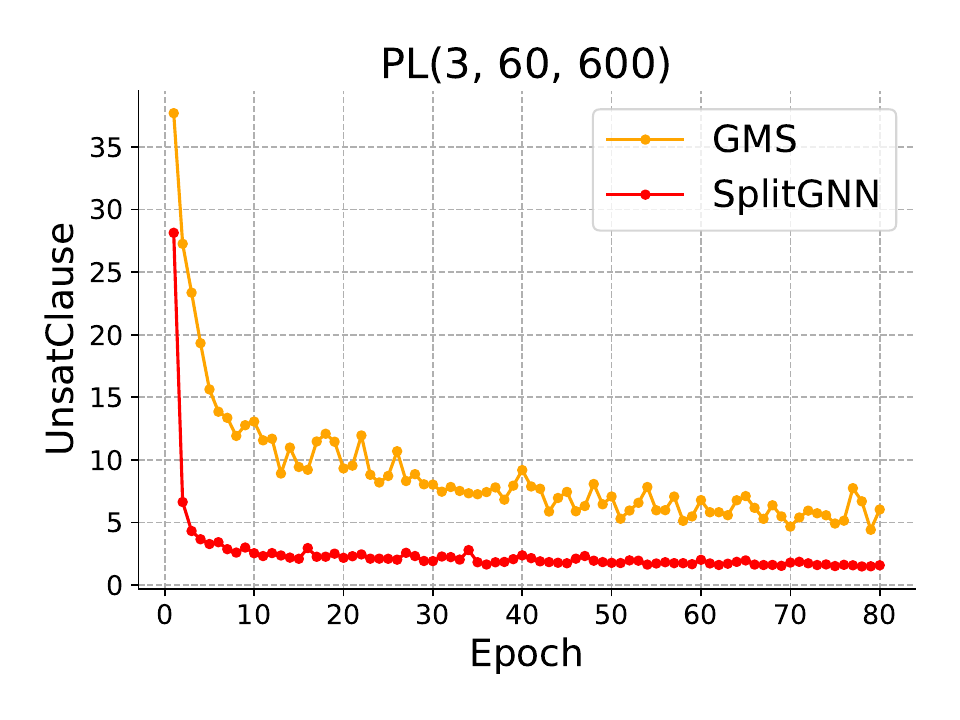}
    \includegraphics[width=0.245\linewidth]{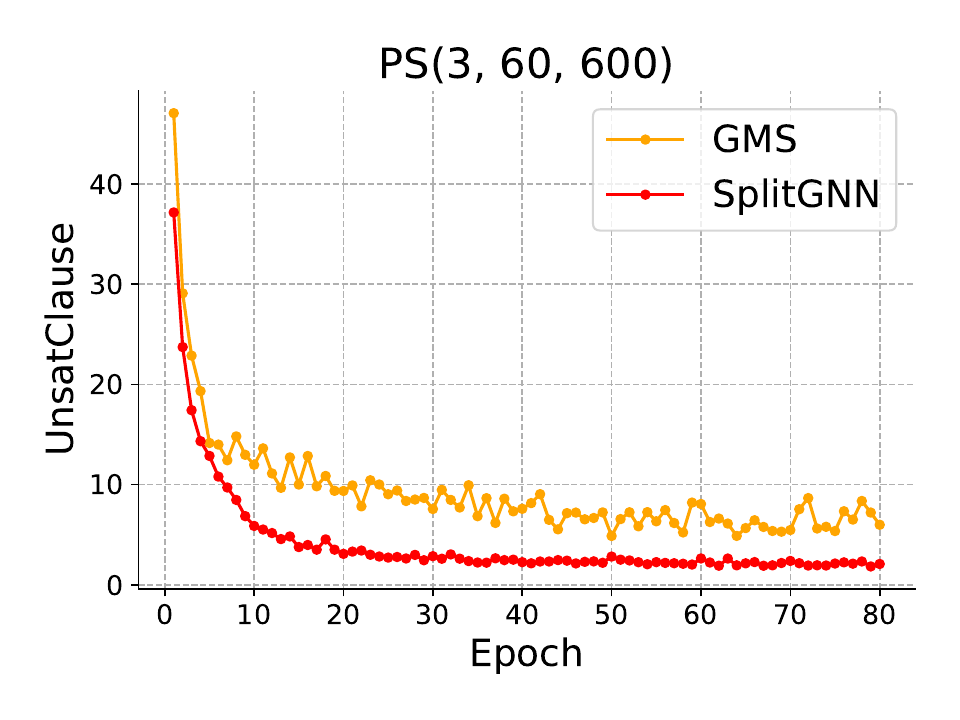}
    \includegraphics[width=0.245\linewidth]{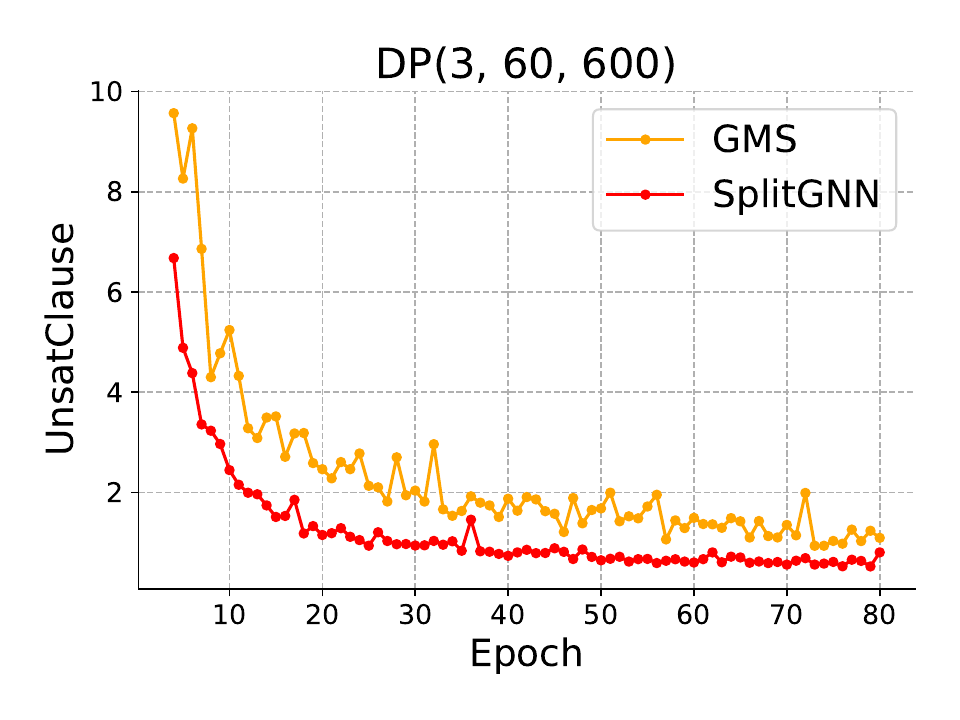}
    \caption{Comparison of the average number of unsatisfied clauses achieved by SplitGNN and GMS-N under different training epochs.}
    \label{fig:unsat-clauses-gms-splitgms}  % 更安全的命名：小写、无空格、用连字符
\end{figure*}

% The experiments demonstrate the improved performance of the modified SplitGNN algorithm on weighted CNF benchmarks. 
We first compare SplitGNN with two baseline GNN models, GMS-N and GMS-E \cite{liu2023can}, on unweighted MaxSAT benchmarks.
Table \ref{tab:Unweighted CNF} shows the average number of unsatisfied clauses of the models testing on different benchmarks.
% Split3GNN and Split4GNN are the models we designed, and the difference between them is whether or not they split the non-tree edges into two kinds. 
SplitGNN\textsuperscript{MP} is a variant of SplitGNN that retains only the message passing component while removing the USB layer.
All models are tested on the benchmarks of the same parameters (i.e., generator, $k$, $n$, and $m$) as those used for training.
The results across four models on unweighted CNF benchmarks highlight the consistent superiority of SplitGNN, which showcase our message passing mechanism based on edge-splitting factor graph contributes to improving learning effectiveness.
% Split4GNN achieves the lowest values in all benchmarks, demonstrating robust generalization over GMS-based methods.
% Notably, Split3GNN also outperforms both GMS variants, particularly in complex scenarios such as PL and PS benchmarks.
% These results underscore the effectiveness of split-level graph architectures in capturing clause-variable dependencies, which likely enhances conflict resolution and reduces unsatisfiable clause assignments. The hierarchical learning framework of SplitGNN models appears to offer a scalable and adaptive solution for MaxSAT challenges.
Moreover, Compared to SplitGNN\textsuperscript{MP}, SplitGNN further improve the quality of solutions across all benchmarks, which demonstrates the positive impact of our USB layer.
For a more detailed presentation of the learning performance of GNN-based models, we also compare the average number of unsatisfied clauses between GMS-N and SplitGNN over different training epochs.
From Figure \ref{fig:unsat-clauses-gms-splitgms}, we observe that SplitGNN exhibits a faster and more stable reduction for the number of unsatisfied clauses, which consistently predicts better solutions than GMS-N under the same training epochs. 
It is worth mentioning that SplitGNN converges approximately $ 3\times$ faster than GMS-N, reaching a lower level of unsatisfied clauses in a short period. This suggests that our SplitGNN model is not only more efficient in terms of convergence speed, but also returns solutions with higher quality when learning to solve unweighted MaxSAT problems.
% This indicates that SplitGNN not only outperforms GMS in terms of convergence speed but also demonstrates better overall performance, providing faster solutions with higher solution quality.
Moreover, even if we reduce the dimensions of hidden layers of SplitGNN to 64, which makes SplitGNN to have significantly fewer parameters than GMS models, it can still achieve comparable results. 
% SplitGNN builds for the MaxSAT problem, and this method has the potential to be used for other graph problems.

\begin{table} [tb]
    \centering
    % \small

    \setlength{\tabcolsep}{0.58mm}
    % \resizebox{\linewidth}{!}{
    \begin{tabular}{lcccc}
    \toprule
    \textbf{Benchmark} & \textbf{{GMS-N}} & \textbf{{GMS-E}} & {\textbf{SplitGNN\textsuperscript{MP}}} & \textbf{{SplitGNN}} \\ \midrule
    UF(2, 60, 600) & 0.517 & 0.573 & 0.371 & \textbf{0.362} \\ %\midrule
    UF(3, 60, 600) & 2.387 & 2.155 & {1.513} & \textbf{1.512} \\ %\midrule
    PL(2, 60, 600) &0.679 & 0.651 & 0.372 & \textbf{0.368} \\ %\midrule
    PL(3, 60, 600) & 2.442 & 2.217 & 1.514 & \textbf{1.446} \\ %\midrule
    % PS(2, 60, 600) & {---} & {---} & {---} & {---} \\ %\midrule
    PS(3, 60, 600) & 2.168 & 2.066 & 1.566 & \textbf{1.507} \\ %\midrule
    DP(3, 60, 600) & 0.593 & 0.504 & 0.353 & \textbf{0.338} \\ 
    \bottomrule
    \end{tabular}
    % }
    \caption{The average number of unsatisfied clauses of GNN-based models on unweighted MaxSAT benchmarks. The results indicate that {SplitGNN} outperforms the two baseline GMS models, and SplitGNN with the USB layer leads to better solutions.}
    \label{tab:Unweighted CNF}
\end{table}

% \resizebox{\linewidth}{!}{
\begin{table}[tb]
\centering

% \sisetup{table-format=3.3} % 全局设置数值格式
% \begin{minipage}[t]{0.48\textwidth} % Train 表格
\centering

\begin{tabular}{@{} l S[table-format=3.3] S[table-format=2.1] @{}} 
% \begin{tabular}{@{} l c c @{}}

    \toprule
    
    %\multicolumn{3}{c}{\textbf{Train}} \\
    % \cmidrule(lr){1-3}
    %\midrule
    \textbf{Benchmark} & \textbf{$\Delta$Obj} & \textbf{VarAcc(\%)} \\
    \midrule
    WUF(2, 60, 600)  &  26.493 & 93.7 \\ 
    WUF(3, 60, 600)  & 134.746 & 81.3 \\ 
    WPL(2, 60, 600)  &  22.642 & 93.6 \\ 
    WPL(3, 60, 600)  & 99.843 & 82.1 \\
    WPS(2, 60, 600)  & 26.330 & 93.7 \\ 
    WPS(3, 60, 600)  & 102.453 & 81.7 \\ 
    WDP(3, 60, 600)  &  19.368 & 94.6 \\ 
    \bottomrule
\end{tabular}

 \caption{The performance of SplitGNN on weighted MaxSAT benchmarks. $\Delta$Obj represents the sum of weights of unsatisfied clauses, and VarAcc represents the variable accuracy with respect to an optimal assignment. While GMS models cannot handle weighted instances, SplitGNN successfully solves these more general instances with high quality.}

% \end{minipage}
\hfill
% \begin{minipage}[t]{0.48\textwidth} % Test 表格

% \end{minipage}
\label{tab:WCNF_Train}
\end{table}
% }

\begin{table*}[htbp]
    \centering
    \begin{tabular}{ccccccc}
        \toprule
        \multirow{2}{*}{\textbf{Benchmark}} & \multirow{2}{*}{\textbf{Loandra}} & \multirow{2}{*}{\textbf{SATLike}} & \multicolumn{2}{c}{\textbf{SplitGNN\textsuperscript{MP}}} & \multicolumn{2}{c}{\textbf{SplitGNN}} \\
        %\cmidrule(lr){4-5} \cmidrule(lr){6-7}
        & & & \textbf{Value} & \textbf{Time (s)} & \textbf{Value} & \textbf{Time (s)}\\
        \midrule
        WUF(2, 2000, 20000)  & 203,318 & 118,768 & 146,474 & 1.95 & \textbf{118,053} & 33.42 \\ 
        WUF(3, 2000, 20000)  & 86,740 & 26,450 & 54,228 & 2.33 & \textbf{25,301} & 34.24\\ 
        WPL(2, 2000, 20000)  & 201,358 & \textbf{130,339} & 138,417 & 1.97 & 130,995 & 38.85 \\ 
        WPL(3, 2000, 20000)  & 83,535 & 32,653 & 51,508 & 2.13 & \textbf{32,395} & 23.87 \\ 
        WPS(2, 2000, 20000)  & 200,316 & \textbf{123,041} & 129,761 & 1.97 & 123,598 & 32.06 \\ 
        WPS(3, 2000, 20000)  & 81,673 & 29,525 & 31,745 & 2.23 & \textbf{28,807} & 39.68\\ 
        WDP(3, 300, 3000)  & 27,928 & 26,911 & 15,170 & 1.10 & \textbf{13,550} & 35.44\\ 
        \bottomrule
    \end{tabular}
         \caption{Comparison of the average sum of weights of unsatisfied clauses ($\Delta$Obj) achieved by SplitGNN and cutting-edge MaxSAT solvers on large and hard testing benchmarks. SplitGNN models are trained on benchmarks with the same distributions but are very small and easy, which demonstrate highly competitive solving abilities on weighted MaxSAT problems.}
 
   % \centering
    % \begin{tabular}{ccccc}
    % \toprule
    % %\multicolumn{5}{c}{\textbf{Test}} \\
    % % \cmidrule(lr){1-4}
    % %\midrule
    % \textbf{Benchmark} & {\textbf{Loandra}} & {\textbf{SATLike}} & {\textbf{SplitGNN\textsuperscript{MP}}} & {\textbf{SplitGNN}} \\
    % \midrule
    % WUF(2, 2000, 20000)  &  {203,318} & {118,768} & {146,474 1.95s} & \textbf{118,053} \\ 
    % WUF(3, 2000, 20000)  &  {86,740} & {26,450} & {54,228 2.33s} & \textbf{25,301} \\ 
    % WPL(2, 2000, 20000)  &  {201,358} & \textbf{130,339} & {138,417 1.97s} & {130,995} \\ 
    % WPL(3, 2000, 20000)  &  {83,535} & {32,653} & {51,508 2.13s} & \textbf{32,395} \\ 
    % WPS(2, 2000, 20000)  &  {200,316} & \textbf{123,041} & {129,761 1.97s} & {123,598} \\ 
    % WPS(3, 2000, 20000)  &  {81,673} & {29,525} & {31,745 2.23s} & \textbf{28,807} \\ 
    % WDP(3, 300, 3000)  &  {27,928} & {26,911} & {15,170 1.10s} & \textbf{13,550} \\ 
    % \bottomrule
    % \end{tabular}
    \label{tab:WCNF_Test}
\end{table*}

In addition, we train and evaluate the performance of SplitGNN on weighted MaxSAT benchmarks.
In Table \ref{tab:WCNF_Train}, we report $\Delta$Obj, which is the sum of weights of unsatisfied clauses, and VarAcc, which is the accuracy of variable assignments that match an optimal solution, of SplitGNN testing on different benchmarks.
% The random weights added to the benchmarks allowed for a fair test of the algorithm’s capability to deal with such weighted problems. 
Note that GMS models are not designed to process weighted MaxSAT problems.
The results indicate that SplitGNN can successfully solve the instances with relatively low $\Delta$Obj values and variable accuracy of over 80\%, which highlights our original contribution on learning to solve these more general and challenging MaxSAT problems with an end-to-end GNN model.

\paragraph*{Results on Large and Hard Benchmarks}
An exciting discovery in existing work is that GNN models trained on small and easy instances from the same distributions can generalize to achieve excellent solving abilities on much larger and more challenging instances \cite{liu2023can}.
We evaluate whether SplitGNN has similar abilities in such scenario on weighted MaxSAT benchmarks in two steps.
% Table \ref{tab:WCNF_Train} and Table \ref{tab:WCNF_Test} presents that we trained SplitGNN on small benchmarks with different distributions, and then used it as initialization to build SplitGNN+US according to the process in Figure \ref{fig:TSMaxSAT}, achieving results similar to or even better than modern MaxSAT solvers Loandra and SATLike. 
% This demonstrates the ability of graph neural networks and search to hybridly solve weighted MaxSAT problems, and provides new ideas for more combinatorial optimization problems.   %%%%%%%%%%%%%%%%%%%%%%
% We build a two-stage optimization framework for SAT solving that demonstrates significant performance improvements. 
First, we train SplitGNN on weighted MaxSAT benchmarks with the 4 generators, $k=\{2,3\}$, $n=60$, and $m=600$, which are very easy instances for current MaxSAT solvers. 
% Next, these trained models are subsequently transferred to initialize our solver on large-scale benchmarks containing 2000 variables and 20,000 clauses. 
Next, for each model trained on a benchmark, we generate a testing benchmark consisting of 2,000 instances, which use the same generator and $k$ as the training set, but $n=2,000$ and $m=20,000$.
Since the instances generated by WDP are harder to solve in practice, we set $k=3$, $n=300$ and $m=3,000$ instead.
These large instances are considered to be challenging even for state-of-the-art MaxSAT solvers.
We evaluate the trained SplitGNN models, and also two cutting-edge heuristic MaxSAT solvers Loandra \cite{berg2020loandra} and SATLike \cite{lei2021satlike}, on the corresponding testing benchmarks. 
The results of average $\Delta$Obj are shown in \ref{tab:WCNF_Test}.
% a novel hybrid methodology that synergistically integrates SplitGNN and stochastic unsurpervised search algorithms with gradient-based backpropagation techniques. 
% Our experimental results reveal that this combined approach achieves faster convergence and higher solution accuracy compared to conventional methods, ultimately surpassing modern SAT solvers in both solution quality and computational efficiency on weighted CNF instances.
SplitGNN produces better solutions for most instances compared with Loandra and SATLike, which indicates that GNN-based methods have great potential in solving challenging real-world MaxSAT problems.
Moreover, SplitGNN solves an instance within 40 seconds, while the solving time limit for Loandra and SATLike are set to 60 seconds. This once again validates the time advantage of GNN-based methods.
Lastly, the comparison between $SplitGNN\textsuperscript{MP}$ and SplitGNN reveals that both the message passing mechanism and the USB layer make contributions to improving the performance.
We provide additional details, including examples, workflows, and experiments, in the appendix.
\section{Conclusion}
% \subsection{Summary} like biconnectivity graph neural network
We present SplitGNN, a novel co-training architecture that combines supervised message passing with hybrid unsupervised solution improvement strategies. 
Edge-splitting factor graph integrates spanning tree to dynamically classify edges into four types: parent edges, child edges, non-tree up edges, and non-tree down edges.
By leveraging these structural distinctions, message passing enhances the ability of SplitGNN to capture critical graph properties, allowing more efficient reasoning. 
The experimental results demonstrate the superior performance of SplitGNN over existing GNN-based approaches, achieving faster convergence and higher accuracy even with reduced parameter dimensions. 
Building on this, SplitGNN integrates hybrid unsupervised strategies, using GPU-accelerated relaxation-based optimization to solve large-scale weighted MaxSAT instances. 
The comparison with state-of-the-art heuristic solvers, particularly on hard industrial-like benchmarks, shows that SplitGNN produces better solutions with minimal computational overhead.
% \subsection{Future Work}

\iffalse
Future work includes extending SplitGNN's edge classification and graph position encoding paradigm to broader NP-hard combinatorial optimization problems, such as graph partitioning and routing optimization, to validate its generalizability. 
Theoretical analysis of SplitGNN with algorithmic principles (e.g., biconnectivity detection) will strengthen its foundational rigor.
% Enhancing scalability through hierarchical decomposition or federated learning could address ultra-large industrial instances. 
Integrating SplitGNN's prediction with symbolic solvers may further balance efficiency and completeness, while energy-efficient adaptations, such as sparsity-aware training, could optimize the resource usage.
\fi
% In addition, dynamic edge reclassification and adaptive relaxation strategies will be investigated to improve robustness against evolving graph structures and heterogeneous data distributions. 
% The combination of neural-symbolic with multimodal, for example, to provide logical reasoning, and security control will also be one of the future challenges. 
% In addition, dynamic edge reclassification and adaptive relaxation strategies will be investigated to improve robustness against evolving graph structures and heterogeneous data distributions. 
% These directions aim to advance neural-symbolic optimization, bridging data-driven learning with classical combinatorial methods.

\bibliography{arxiv}

\clearpage
\appendix

% \section{Appendix} 

\section*{Frameworks of Models}

\begin{algorithm}[htbp]
    \caption{Unsupervised Solution Boosting}
    \label{alg:TSMaxSAT}
    \raggedright % 强制左对齐
    \textbf{Input}: Matrix $W_{unit}$ and $U$ from CNF, vector $S$ where $S_j = -|C_j|$, clauses weight $CW$\\
    \textbf{Parameter}: Time limit ($\mathit{TLimit}$), learning rate ($\gamma$), length ($len$), size ($k$)\\
    \textbf{Output}: The best found solution and its cost\\%, or “No solution found”.
    \begin{algorithmic}[1] %[1] enables line numbers
        \WHILE{solving time $< \mathit{TLimit}$}
        \STATE Initialize $x$ with the output of message passing;
        \STATE $L \leftarrow len$;
        \FOR{$\mathit{step} = 0$ to $L$}
            \STATE Calculate $\alpha$ where $\alpha_i = 1$ \textbf{if} $x_i > 0$, \textbf{else} 0;
            \STATE Calculate ${cost}(\alpha) = (\alpha * W == S) \cdot CW$;
            \IF {$\alpha$ is feasible \textbf{and} $cost^* > {cost}(\alpha)$}
            \STATE {$\alpha^* \leftarrow \alpha; \ \text{cost}^* \leftarrow \text{cost}(\alpha); \ L \leftarrow \mathit{step} + len;$}
                \IF{$\text{cost}^* == 0$}
                \STATE \textbf{return} $ \alpha^*$, $\text{cost}^*;$
                \ENDIF
            \ENDIF
            \STATE score $:=$ Score Calculation with Sparse Matrix;
            \IF{$(D := \{x \mid \text{score}(x) > 0\}) \neq \emptyset$}
                \STATE $v :=$ Select the minimum $k$ variables in $D$;
                %\STATE $v :=$ The selected variable with the maximum score
                \STATE $x_v \leftarrow -x_v$;
                \ELSE
                    \STATE Run forward pass, $f(x) = \tau * \frac{\tanh(x)}{score^2+\epsilon}  \times W_{unit}$;
                    % \STATE Calculate loss $L$, for $f(x)$ w.r.t variables contributing to the unsatisfied clauses, U.
                    \STATE $\mathit{loss}$ = MSE($\frac{f[U] \circ CW}{S[U]}$, zeros\_like$(U)$);
                    % \IF {$\exists$ falsified hard clauses,}
                    %     \STATE loss = MSE($f[HU] \circ HCW$, zeros\_like$(HU)$)
                    % \ELSE
                        % \STATE loss = MSE($f[SU] \circ SCW$, zeros\_like$(SU)$)
                    % \ENDIF
                    \STATE Run backpropagation with $\mathit{loss}$;
            \ENDIF
        \ENDFOR
        \ENDWHILE
        \STATE \textbf{return} $\alpha^*$, $\text{cost}^*$;
        % \IF{$\alpha^* \neq \emptyset$}
        %     \STATE \textbf{return} $\alpha^*$ and $\text{cost}^*$;
        % \ELSE
        %     \STATE \textbf{return} No solution found;
        % \ENDIF
    \end{algorithmic}
\end{algorithm}

\section{Example} 
Consider the following MaxSAT problem in CNF:
\[ \mathcal{F} = (\neg x_1) \land (\neg x_2) \land (x_1 \lor x_2). \]
The clause weights are 3, 4, and 5, respectively. 
This formula is not satisfiable, and it is evident that any combination of binary assignments for \( x_1 \) and \( x_2 \) can satisfy, at most, two clauses.
Assume that \( \mathbf{x} = [x_1, x_2] \) is initialized randomly as \([-0.79, 1.34]\), which is interpreted as \( x_1 = 0 \) and \( x_2 = 1 \). 
The output of the forward pass is: 
\[ U = \left( [-1, 1] \cdot  \begin{bmatrix} -1 & 0 & 1 \\ 0 & -1 & 1 \end{bmatrix} == [-1, -1, -2] \right), \]
\[O = \left( [-1, 1] \cdot  \begin{bmatrix} -1 & 0 & 1 \\ 0 & -1 & 1 \end{bmatrix} == [1, 1, 0] \right). \]
Therefore, \(U = [0, 1, 0], \ \  O = [1, 0, 1]\).
\[M_u = [0, 1, 0] \circ \begin{bmatrix} 3 & 0 & 5 \\ 0 & 4 & 5 \end{bmatrix}=\begin{bmatrix} 0 & 0 & 0 \\ 0 & 4 & 0 \end{bmatrix},\]
\[M_0 = [1, 0, 1] \circ \begin{bmatrix} -3 & 0 & 5 \\ 0 & -4 & 5 \end{bmatrix} \circ \begin{bmatrix} -1 \\ 1\end{bmatrix}=\begin{bmatrix} 3 & 0 & -5 \\ 0 & 0 & 5 \end{bmatrix},\]
\[max\_pos = \begin{bmatrix}0 \\ 0 \\ 1\end{bmatrix}, \ \ max\_val = \begin{bmatrix}3 \\ 0 \\ 5\end{bmatrix},\]
\[Add = \begin{bmatrix} 0 \\ 4\end{bmatrix}, \ \ Del = \begin{bmatrix} 3 \\ 5\end{bmatrix},\]
\[Score = \begin{bmatrix} -3 \\ -1\end{bmatrix},\]
\[ f(\mathbf{x}) = 0.5 * \frac{\tanh \left( \begin{bmatrix} -0.79 & 1.34 \end{bmatrix} \right)}{[(-3)^2+0.01, (-1)^2+0.01]} \cdot \begin{bmatrix} -1 & 0 & 1 \\ 0 & -1 & 1 \end{bmatrix} \]
\[ f(\mathbf{x}) = \begin{bmatrix} -0.04 & -0.43 & 0.47 \end{bmatrix} \]
The process of finding the optimal solution is $(0, 1) \xrightarrow{relex} (0, 0) \xrightarrow{score} (1, 0)$.

% Due to space constraints, we list the size (\#vars/\#clauses) of the three largest instances in \autoref{tab:complete_problem_stats}. 

% \begin{table}[!htbp]
% \centering
% \caption{Part statistics of Boolean variables and clauses across all 50 problem instances (CB, GT, PAR, PHP datasets)}
% % \begin{tabular} {@{} l *{8}{S[table-format=3.3]} @{}} 
% \resizebox{1.0\linewidth}{!}{
% \begin{tabular}{ccc|cc|cc|cc}
% \toprule
% {} & \multicolumn{2}{c}{\textbf{CB}} & \multicolumn{2}{c}{\textbf{GT}} & \multicolumn{2}{c}{\textbf{PAR}} & \multicolumn{2}{c}{\textbf{PHP}} \\
% {} & {\#vars} & {\#clauses} & {\#vars} & {\#clauses} & {\#vars} & {\#clauses} & {\#vars} & {\#clauses} \\
% \midrule 
% {48}  & {19008}  & {33026}   & {2352}   & {111769}  & {4656}   & {442417}  & {2352}   & {56497}   \\
% {49}  & {19796}  & {34400}   & {2450}   & {118875}  & {4851}   & {470646}  & {2450}   & {60075}   \\
% {50}  & {20600}  & {35802}   & {2550}   & {126276}  & {5050}   & {500051}  & {2550}   & {63801}   \\
% \bottomrule
% \end{tabular}
% }
% \label{tab:complete_problem_stats}

% \end{table} 

\section{Generalizing to Structural Benchmarks}

\begin{table*}[tb]
    \centering
    % \resizebox{1.0\linewidth}{!}{
    \begin{tabular}{ccccccc}
        \toprule
        \textbf{Benchmark} & \textbf{Average size} & \textbf{Max size} & \textbf{{SATLike}} & \textbf{{Loandra}} & \textbf{{torchmSAT}} & \textbf{{SplitGNN}} \\
        \midrule
        PHP & {1,275 / 31,950.5} & {2,550 / 63,801} & 4  & \textbf{0}  & 22 & \textbf{0} \\
        PAR & {2,525 / 250,025.5} & {5,050 / 500,051} & 2   & 5          & 74 & \textbf{0} \\
        GT  & {1,275 / 6,313.5} & {2,550 / 126,276} & 20  & \textbf{0}  & 53 & \textbf{0} \\
        CB  & {10,300 / 17,901} & {20,600 / 35,802} & 233  & \textbf{0}  & 1076 &  \textbf{0} \\
        % WPHP & 4  & \textbf{0}  & 22 & \textbf{0} \\
        % WPAR & 2   & 5          & 74 & \textbf{0} \\
        % WGT  & 20  & \textbf{0}  & 53 & \textbf{0} \\
        % WCB  & 233  & \textbf{0}  & 1076 &  \textbf{0} \\
        \bottomrule
    \end{tabular} 
    % }
    \caption{
        Comparison of the maximum regret on structural testing benchmarks.
        SplitGNN generalizes across distributions and predicts the best solution among the four solvers on all instances.
        % The regret $\text{regret}(s, i)$ of a solver $s$ on instance $i$ is the difference between the cost of the best solution found by \( s \) and the cost of the best known solution:
        % $\text{regret}(s, i) = \text{cost}_{s,i} - \text{cost}_{\text{best},i}.$     
        % Cost is defined as the number of unsatisfied clauses. 
        % Considering the result of maxhs solver is the optimal solution (i.e., $\text{regret} = 0$), the table presents the average regret of SplitGNN+US as compared to SATLike, Loandra and torchmSAT. 
        % In SplitGNN+US, regret decreases as the solving time limit increases.
    }
    \label{tab:regret_comparison}
   
\end{table*}

We further evaluate the generalization ability of SplitGNN on the benchmarks from different distributions.
To get new testing instances, we utilize PySAT to generate 4 representative structural MaxSAT benchmarks, each embodying combinatorial principles extensively studied within the realm of propositional proof complexity: the pigeonhole principle (PHP), the parity principle (PAR), the greater-than ordering principle (GT), and the mutilated chessboard principle (CB).
Each benchmark contains 50 instances.
After training SplitGNN on UF(3,60,600) benchmark, we test SplitGNN with baselines Loandra, SATLike, and torchmSAT\footnote{torchmSAT has no public code, and we reproduce it with much better performance than that in torchmsat.}. 
torchmSAT is a GPU-accelerated approximate MaxSAT solver but it is not GNN-based, so it cannot learn from instances with message passing.
% , achieves significantly lower regret values, particularly in the SplitGNN+US solver. 
% For example, on the CB dataset, SplitGNN+US achieves the lowest average regret, while the other solvers show much higher values, indicating that SplitGNN is highly effective in minimizing the number of unsatisfied clauses and finding solutions closer to the optimal.
\ref{tab:regret_comparison} shows the performance of these methods evaluated using the maximum regret. For a solver $S$ and a benchmark $I$, $\text{max\_regret}(S,I)=\max_{i \in I}{\text{regret}(S,i)}$, where $\text{regret}(S,i)$ is the difference of $\Delta Obj$ between the solution solved by $S$ and the best known solution on instance $i$.
Maximum regret equal to 0 means that a solver finds the best known solution on all instances.
From the results, SplitGNN stands out and it is strongly robust across all benchmarks.
More specifically, the comparison with torchmSAT confirms that the learning capacity of GNNs is indispensable in order to achieve superior performance.
Furthermore, we have attempted to evaluate SplitGNN on standard MaxSAT Evaluation benchmarks. 
The instances in MaxSAT Evaluations are partial MaxSAT problems. 
We evaluated SplitGNN on 70 instances from MSE2023 (auctions, synplicate, pseudoBoolean, and planning). 
It satisfied all hard clauses and many soft clauses by setting their weights to the sum of soft clauses weight. % ($10^{12}$). 
% For partial MaxSAT, the integration of traditional search algorithms could be a better solution for balancing hard and soft clauses, which we leave for future work.

% In the PHP dataset, SplitGNN+US again stands out with an average regret of 0, indicating that it found the best solution. 
% These results demonstrate the robustness of SplitGNN+US across a variety of benchmark datasets, demonstrating its ability to find high-quality solutions with minimal computational effort. 
% Overall, the experimental results reinforce the superior performance of SplitGNN, with its ability to achieve near-optimal solutions in a variety of instances, particularly when comparing it against other solvers in terms of regret.
% Except for the CB problem where we add some small data from that distribution, we can train on UF(3,60,600) and then migrate to other distributions to get the optimal solution by SplitGNN+US.
% The lager CNF with these distributions of these migrations all have thousands of variables and tens of thousands of clauses. 
% Our code and models will be open source.

\section*{Model Implementation and Training} 

\begin{table}[htb]
    \centering
    \vspace{5pt}
    \begin{tabular}{lcc}
        \toprule
        \textbf{Model}  & \textbf{\#Dim} & \textbf{\#Params} \\
        \midrule
        GMS\_N      & 128   & 478,849     \\
        GMS\_E      & 128   & 496,001     \\
        SplitGNN  &\ \ 80   & 446,561     \\
        \bottomrule
    \end{tabular}
    \caption{Comparison of the number of dimensions and parameters of the GNN models.}
    \label{tab:plain}
\end{table}

% We implement SplitGNN in Python using the PyTorch library\footnote{For reproducibility, our code and models will be publicly available after acceptance.} \cite{paszkepytorch}.
We implement SplitGNN in Python using the PyTorch library. % \cite{paszkepytorch}.
For the projection head, we use multi-layer perceptrons (MLPs) with one hidden layer, where the dimension of hidden layer is set to 80.
The number of GNN layers is set to 30.
As a key baseline method, We download the implementation of the GMS models and preserve their default settings\footnote{\url{https://github.com/minghao-liu/GMS}}.
For a fair comparison, we summarize the dimensions of hidden layers and the number of parameters for all models in table \ref{tab:plain}.
It can be observed that the total parameter number of SplitGNN is lower than that of the GMS models.

We train all GNN models using the AdamW optimizer with learning rate $2 \times 10^{-5}$ and weight decay $10^{-10}$. 
The batch size is 256 and the maximum training epochs are 300.
Temperature coefficient $\tau$ is 0.5 and epsilon $\epsilon$ is 0.01. 
We use dropout to enhance the accuracy of the models and prevent overfitting. 
All models are trained and tested on a server with Intel Core i7-14700F CPU, NVIDIA GEFORCE 4060 Ti GPU, Ubuntu 24.04 OS, and 32GB memory.

\newpage
\makeatletter
\@ifundefined{isChecklistMainFile}{
  % We are compiling a standalone document
  \newif\ifreproStandalone
  \reproStandalonetrue
}{
  % We are being \input into the main paper
  \newif\ifreproStandalone
  \reproStandalonefalse
}
\makeatother

\ifreproStandalone
\documentclass[letterpaper]{article}
\usepackage[submission]{aaai2026}
\setlength{\pdfpagewidth}{8.5in}
\setlength{\pdfpageheight}{11in}
\usepackage{times}
\usepackage{helvet}
\usepackage{courier}
\usepackage{xcolor}
\frenchspacing

% \begin{document}
\fi
\setlength{\leftmargini}{20pt}
\makeatletter\def\@listi{\leftmargin\leftmargini \topsep .5em \parsep .5em \itemsep .5em}
\def\@listii{\leftmargin\leftmarginii \labelwidth\leftmarginii \advance\labelwidth-\labelsep \topsep .4em \parsep .4em \itemsep .4em}
\def\@listiii{\leftmargin\leftmarginiii \labelwidth\leftmarginiii \advance\labelwidth-\labelsep \topsep .4em \parsep .4em \itemsep .4em}\makeatother

\setcounter{secnumdepth}{0}
\renewcommand\thesubsection{\arabic{subsection}}
\renewcommand\labelenumi{\thesubsection.\arabic{enumi}}

\newcounter{checksubsection}
\newcounter{checkitem}[checksubsection]

\newcommand{\checksubsection}[1]{%
  \refstepcounter{checksubsection}%
  \paragraph{\arabic{checksubsection}. #1}%
  \setcounter{checkitem}{0}%
}

\newcommand{\checkitem}{%
  \refstepcounter{checkitem}%
  \item[\arabic{checksubsection}.\arabic{checkitem}.]%
}
\newcommand{\question}[2]{\normalcolor\checkitem #1 #2 \color{blue}}
\newcommand{\ifyespoints}[1]{\makebox[0pt][l]{\hspace{-15pt}\normalcolor #1}}

\section*{Reproducibility Checklist}

\vspace{1em}
\hrule
\vspace{1em}

\textbf{Instructions for Authors:}

This document outlines key aspects for assessing reproducibility. Please provide your input by editing this \texttt{.tex} file directly.

For each question (that applies), replace the ``Type your response here'' text with your answer.

\vspace{1em}
\noindent
\textbf{Example:} If a question appears as
\begin{center}
\noindent
\begin{minipage}{.9\linewidth}
\ttfamily\raggedright
\string\question \{Proofs of all novel claims are included\} \{(yes/partial/no)\} \\
Type your response here
\end{minipage}
\end{center}
you would change it to:
\begin{center}
\noindent
\begin{minipage}{.9\linewidth}
\ttfamily\raggedright
\string\question \{Proofs of all novel claims are included\} \{(yes/partial/no)\} \\
yes
\end{minipage}
\end{center}
Please make sure to:
\begin{itemize}\setlength{\itemsep}{.1em}
\item Replace ONLY the ``Type your response here'' text and nothing else.
\item Use one of the options listed for that question (e.g., \textbf{yes}, \textbf{no}, \textbf{partial}, or \textbf{NA}).
\item \textbf{Not} modify any other part of the \texttt{\string\question} command or any other lines in this document.\\
\end{itemize}

You can \texttt{\string\input} this .tex file right before \texttt{\string\end\{document\}} of your main file or compile it as a stand-alone document. Check the instructions on your conference's website to see if you will be asked to provide this checklist with your paper or separately.

\vspace{1em}
\hrule
\vspace{1em}

% The questions start here

\checksubsection{General Paper Structure}
\begin{itemize}

\question{Includes a conceptual outline and/or pseudocode description of AI methods introduced}{(yes/partial/no/NA)}
yes

\question{Clearly delineates statements that are opinions, hypothesis, and speculation from objective facts and results}{(yes/no)}
yes

\question{Provides well-marked pedagogical references for less-familiar readers to gain background necessary to replicate the paper}{(yes/no)}
yes

\end{itemize}
\checksubsection{Theoretical Contributions}
\begin{itemize}

\question{Does this paper make theoretical contributions?}{(yes/no)}
yes

	\ifyespoints{\vspace{1.2em}If yes, please address the following points:}
        \begin{itemize}
	
	\question{All assumptions and restrictions are stated clearly and formally}{(yes/partial/no)}
	yes

	\question{All novel claims are stated formally (e.g., in theorem statements)}{(yes/partial/no)}
	yes

	\question{Proofs of all novel claims are included}{(yes/partial/no)}
	yes

	\question{Proof sketches or intuitions are given for complex and/or novel results}{(yes/partial/no)}
	yes

	\question{Appropriate citations to theoretical tools used are given}{(yes/partial/no)}
	yes

	\question{All theoretical claims are demonstrated empirically to hold}{(yes/partial/no/NA)}
	yes

	\question{All experimental code used to eliminate or disprove claims is included}{(yes/no/NA)}
	yes
	
	\end{itemize}
\end{itemize}

\checksubsection{Dataset Usage}
\begin{itemize}

\question{Does this paper rely on one or more datasets?}{(yes/no)}
yes

\ifyespoints{If yes, please address the following points:}
\begin{itemize}

	\question{A motivation is given for why the experiments are conducted on the selected datasets}{(yes/partial/no/NA)}
	yes

	\question{All novel datasets introduced in this paper are included in a data appendix}{(yes/partial/no/NA)}
	yes

	\question{All novel datasets introduced in this paper will be made publicly available upon publication of the paper with a license that allows free usage for research purposes}{(yes/partial/no/NA)}
	yes

	\question{All datasets drawn from the existing literature (potentially including authors' own previously published work) are accompanied by appropriate citations}{(yes/no/NA)}
	yes

	\question{All datasets drawn from the existing literature (potentially including authors' own previously published work) are publicly available}{(yes/partial/no/NA)}
	yes

	\question{All datasets that are not publicly available are described in detail, with explanation why publicly available alternatives are not scientifically satisficing}{(yes/partial/no/NA)}
	yes

\end{itemize}
\end{itemize}

\checksubsection{Computational Experiments}
\begin{itemize}

\question{Does this paper include computational experiments?}{(yes/no)}
yes

\ifyespoints{If yes, please address the following points:}
\begin{itemize}

	\question{This paper states the number and range of values tried per (hyper-) parameter during development of the paper, along with the criterion used for selecting the final parameter setting}{(yes/partial/no/NA)}
	yes

	\question{Any code required for pre-processing data is included in the appendix}{(yes/partial/no)}
	yes

	\question{All source code required for conducting and analyzing the experiments is included in a code appendix}{(yes/partial/no)}
	yes

	\question{All source code required for conducting and analyzing the experiments will be made publicly available upon publication of the paper with a license that allows free usage for research purposes}{(yes/partial/no)}
	yes
        
	\question{All source code implementing new methods have comments detailing the implementation, with references to the paper where each step comes from}{(yes/partial/no)}
	yes

	\question{If an algorithm depends on randomness, then the method used for setting seeds is described in a way sufficient to allow replication of results}{(yes/partial/no/NA)}
	yes

	\question{This paper specifies the computing infrastructure used for running experiments (hardware and software), including GPU/CPU models; amount of memory; operating system; names and versions of relevant software libraries and frameworks}{(yes/partial/no)}
	yes

	\question{This paper formally describes evaluation metrics used and explains the motivation for choosing these metrics}{(yes/partial/no)}
	yes

	\question{This paper states the number of algorithm runs used to compute each reported result}{(yes/no)}
	yes

	\question{Analysis of experiments goes beyond single-dimensional summaries of performance (e.g., average; median) to include measures of variation, confidence, or other distributional information}{(yes/no)}
	yes

	\question{The significance of any improvement or decrease in performance is judged using appropriate statistical tests (e.g., Wilcoxon signed-rank)}{(yes/partial/no)}
	yes

	\question{This paper lists all final (hyper-)parameters used for each model/algorithm in the paper’s experiments}{(yes/partial/no/NA)}
	yes

\end{itemize}
\end{itemize}
\ifreproStandalone
% \end{document}
\fi

\end{document}